\documentclass[11pt,a4paper]{article}
\usepackage{amsmath}
\usepackage[hyperref]{emnlp2020}
\usepackage{times}
\usepackage{latexsym}
\usepackage{amssymb}
\usepackage{breqn}
\usepackage{graphicx}
\usepackage{enumitem}
\usepackage{anyfontsize}

\usepackage{booktabs}  
\usepackage{scalerel,xparse}

\usepackage{microtype}
\usepackage{arydshln}
\aclfinalcopy 

\makeatletter
\def\thickhline{%
  \noalign{\ifnum0=`}\fi\hrule \@height \thickarrayrulewidth \futurelet
   \reserved@a\@xthickhline}
\def\@xthickhline{\ifx\reserved@a\thickhline
               \vskip\doublerulesep
               \vskip-\thickarrayrulewidth
             \fi
      \ifnum0=`{\fi}}
\makeatother

\newlength{\thickarrayrulewidth}
\title{Claim Check-Worthiness Detection as Positive Unlabelled Learning}

\author{Dustin Wright \and Isabelle Augenstein \\
  Dept. of Computer Science \\
  University of Copenhagen \\
  Denmark \\
  \texttt{\{dw|augenstein\}@di.ku.dk}}

\date{}

\begin{document}
\maketitle
\begin{abstract}

As the first step of automatic fact checking, claim check-worthiness detection is a critical component of fact checking systems. There are multiple lines of research which study this problem: check-worthiness ranking from political speeches and debates, rumour detection on Twitter, and citation needed detection from Wikipedia. 
To date, there has been no structured comparison of these various tasks to understand their relatedness, and no investigation into whether or not a unified approach to all of them is achievable.
In this work, we illuminate a central challenge in claim check-worthiness detection underlying all of these tasks, being that they hinge upon detecting both how factual a sentence is, as well as how likely a sentence is to be believed without verification. As such, annotators only mark those instances they judge to be clear-cut check-worthy. Our best performing method is a unified approach which automatically corrects for this using a variant of positive unlabelled learning that finds instances which were incorrectly labelled as not check-worthy.
In applying this, we outperform the state of the art in two of the three tasks studied for claim check-worthiness detection in English.

\end{abstract}

\section{Introduction}
Misinformation is being spread online at ever increasing rates \cite{del2016spreading} and has been identified as one of society's most pressing issues by the World Economic Forum~\cite{howell2013digital}. In response, there has been a large increase in the number of organizations performing fact checking \cite{graves2016rise}. However, the rate at which misinformation is introduced and spread vastly outpaces the ability of any organization to perform fact checking, so only the most salient claims are checked. This obviates the need for being able to automatically find check-worthy content online and verify it.

\begin{figure}[t]
  
  \centering
    \includegraphics[width=0.5\textwidth]{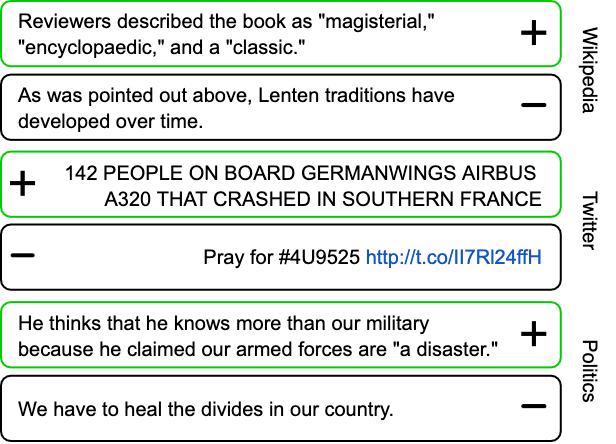}
    \caption{Examples of check-worthy and non check-worthy statements from three different domains. Check-worthy statements are those which were judged to require evidence or a fact check.}
    \label{fig:check-worthy-examples}
\end{figure}
The natural language processing and machine learning communities have recently begun to address the problem of automatic fact checking~\cite{vlachos2014fact,hassan2017claimbuster, thorne-vlachos-2018-automated-fixed,conf/emnlp/AugensteinLWLHH19,conf/acl/AtanasovaSLA20,atanasova2020generating,ostrowski2020multihop,allein2020timeaware}. The first step of automatic fact checking is claim check-worthiness detection, a text classification problem where, given a statement, one must predict if the content of that statement makes ``an assertion about the world that is checkable'' \cite{konstantinovskiy2018towards}.
There are multiple isolated lines of research which have studied variations of this problem. \autoref{fig:check-worthy-examples} provides examples from three tasks which are studied in this work: rumour detection on Twitter~\cite{zubiaga2016analysing,journals/ipm/ZubiagaKLPLBCA18}, check-worthiness ranking in political debates and speeches~\cite{atanasova2018overview,elsayed2019overview,barron2020checkthat}, and citation needed detection on Wikipedia~\cite{redi2019citation}. Each task is concerned with a shared underlying problem: detecting claims which warrant further verification. However, no work has been done to compare all three tasks to understand shared challenges in order to derive shared solutions, which could enable improving claim check-worthiness detection systems across multiple domains. 



Therefore, we ask the following main research question in this work: are these all variants of the same task, and if so, is it possible to have a unified approach to all of them? We answer this question by investigating the problem of annotator subjectivity, where annotator background and expertise causes their judgement of what is check-worthy to differ, leading to false negatives in the data~\cite{konstantinovskiy2018towards}. Our proposed solution is \textit{Positive Unlabelled Conversion (PUC)}, an extension of Positive Unlabelled (PU) learning, which converts negative instances into positive ones based on the estimated prior probability of an example being positive. We demonstrate that a model trained using \textit{PUC} improves performance on English \textit{citation needed detection} and 
\textit{Twitter rumour detection}. We also show that by pretraining a model on citation needed detection, one can further improve results on Twitter rumour detection over a model trained solely on rumours, highlighting that a unified approach to these problems is achievable. Additionally, we show that one attains better results on 
\textit{political speeches}  check-worthiness ranking without using any form of PU learning, arguing through a dataset analysis that the labels are much more subjective than the other two tasks.


The \textbf{contributions} of this work are as follows:
\begin{enumerate}[noitemsep]
    \item The first thorough comparison of multiple claim check-worthiness detection tasks.
    \item \textit{Positive Unlabelled Conversion (PUC)}, a novel extension of PU learning to support check-worthiness detection across domains.
    \item Results demonstrating that a unified approach to check-worthiness detection is achievable for 2 out of 3 tasks, improving over the state-of-the-art for those tasks.
\end{enumerate}
\section{Related Work}

\subsection{Claim Check-Worthiness Detection}
As the first step in automatic fact checking, claim check-worthiness detection is a binary classification problem which involves determining if a piece of text makes ``an assertion about the world which can be checked''~\cite{konstantinovskiy2018towards}. We adopt this broad definition as it allows us to perform a structured comparison of many publicly available datasets. The wide applicability of the definition also allows us to study if and how a unified cross-domain approach could be developed. 

Claim check-worthiness detection can be subdivided into three distinct domains: rumour detection on Twitter, check-worthiness ranking in political speeches and debates, and citation needed detection on Wikipedia. A few studies have been done which attempt to create full systems for mining check-worthy statements, including the works of \citet{konstantinovskiy2018towards}, ClaimRank \cite{jaradat2018claimrank}, and ClaimBuster \cite{hassan2017claimbuster}. They develop full software systems consisting of relevant source material retrieval, check-worthiness classification, and dissemination to the public via end-user applications. These works are focused solely on the political domain, using data from political TV shows, speeches, and debates. In contrast, in this work we study the claim check-worthiness detection problem across three domains which have publicly available data: Twitter~\cite{zubiaga2017exploiting}, political speeches~\cite{atanasova2018overview}, and Wikipedia~\cite{redi2019citation}.


\paragraph{Rumour Detection on Twitter}
Rumour detection on Twitter is primarily studied using the PHEME dataset~\cite{zubiaga2016analysing}, a set of tweets and associated threads from breaking news events which are either rumourous or not. Published systems which perform well on this task include contextual models (e.g. conditional random fields) acting on a tweet's thread~\cite{zubiaga2017exploiting,journals/ipm/ZubiagaKLPLBCA18}, identifying salient rumour-related words~\cite{abulaish2019graph}, and using a GAN to generate misinformation in order to improve a downstream discriminator~\cite{ma2019detect}.

\paragraph{Political Speeches}
For political speeches, the most studied datasets come from the Clef CheckThat! shared tasks ~\cite{atanasova2018overview, elsayed2019overview,barron2020checkthat} and ClaimRank~\cite{jaradat2018claimrank}. The data consist of transcripts of political debates and speeches where each sentence has been annotated by an independent news or fact-checking organization for whether or not the statement should be checked for veracity. The most recent and best performing system on the data considered in this paper consists of a two-layer bidirectional GRU network which acts on both word embeddings and syntactic parse tags~\cite{hansen2019neural}. In addition, they augment the native dataset with weak supervision from unlabelled political speeches.

\paragraph{Citation Needed Detection}
Wikipedia citation needed detection has been investigated recently in~\cite{redi2019citation}. The authors present a dataset of sentences from Wikipedia labelled for whether or not they have a citation attached to them. They also released a set of sentences which have been flagged as not having a citation but needing one (i.e. \textit{unverified}). In contrast to other check-worthiness detection domains, there are much more training data available on Wikipedia. However, the rules for what requires a citation do not necessarily capture all ``checkable'' statements, as ``all material in Wikipedia articles must be verifiable''~\cite{redi2019citation}. 
Given this, we view Wikipedia citation data as a set of positive and unlabelled data: statements which have attached citations are positive samples of check-worthy statements, and within the set of statements without citations there exist some positive samples (those needing a citation) and some negative samples. 
Based on this, this domain constitutes the most general formulation of check-worthiness among the domains we consider. Therefore, we experiment with using data from this domain as a source for transfer learning, training variants of PU learning models on it, then applying them to target data from other domains.

\subsection{Positive Unlabelled Learning}
PU learning methods attempt to learn good binary classifiers given only positive labelled and unlabelled data. Recent applications where PU learning has been shown to be beneficial include detecting deceptive reviews online~\cite{li2014spotting,ren2014positive}, keyphrase extraction~\cite{sterckx2016supervised} and named entity recognition~\cite{peng2019distantly}. For a survey on PU learning, see~\cite{bekker2018learning}, and for a formal definition of PU learning, see \S\ref{sec:pu_learning}.

Methods for learning positive-negative (PN) classifiers from PU data have a long history~\cite{denis1998pac,de1999positive,letouzey2000learning}, with one of the most seminal papers being from~\citet{elkan2008learning}. In this work, the authors show that by assuming the labelled samples are a random subset of all positive samples, one can utilize a classifier trained on PU data in order to train a different classifier to predict if a sample is positive or negative. The process involves training a PN classifier with positive samples being shown to the classifier once and \textit{unlabelled} samples shown as \textit{both} a positive sample and a negative sample. The loss for the duplicated samples is weighted by the confidence of a PU classifier that the sample is positive.

Building on this, du Plessis et al.~\shortcite{du2014analysis} propose an unbiased estimator which improves the estimator introduced in~\cite{elkan2008learning} by balancing the loss for positive and negative classes. The work of Kiryo et al.~\shortcite{kiryo2017positive} extends this method to improve the performance of deep networks on PU learning. Our work builds on the method of Elkan and Noto~\shortcite{elkan2008learning} by relabelling samples which are highly confidently positive.

\section{Methods}
\begin{figure*}[t]
  \centering
    \includegraphics[width=0.95\textwidth]{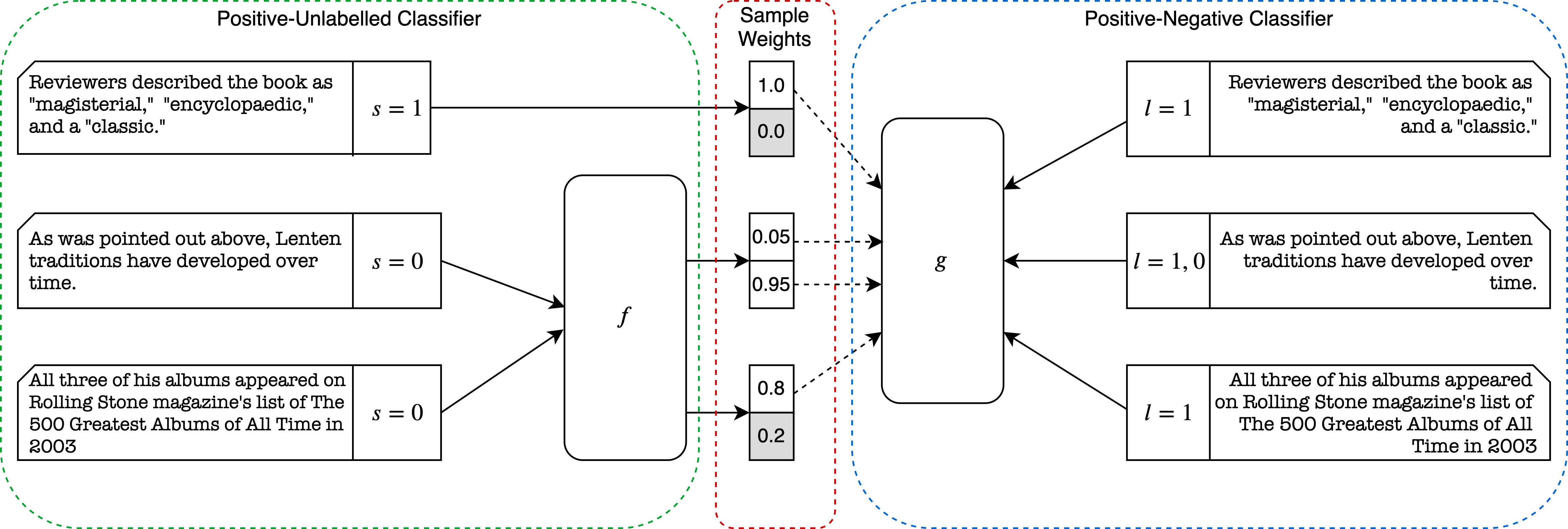}
    \caption{High level view of \textit{PUC}. A PU classifier ($f$, green box) is first learned using PU data (with $s$ indicating if the sample is positive or unlabelled). From this the prior probability of a sample being positive is estimated. Unlabelled samples are then ranked by $f$ (red box) and the most positive samples are converted into positives until the dataset is balanced according to the estimated prior. The model $g$ is then trained using the duplication and weighting method of \citet{elkan2008learning} as described in \S\ref{sec:pu_learning} with labels $l$ (blue box). Greyed out boxes are negative weights which are ignored when training the classifier $g$, as those examples are only trained as positives.}
      \label{fig:puc}
\end{figure*}
The task considered in this paper is to predict if a statement makes ``an assertion about the world that is checkable'' \cite{konstantinovskiy2018towards}. 
As the subjectivity of annotations for existing data on claim check-worthiness detection is a known problem~\cite{konstantinovskiy2018towards}, we view the data as a set of positive and unlabelled (PU) data. In addition, we unify our approach to each of them by viewing Wikipedia data as an abundant source corpus.
Models are then trained on this source corpus using variants of PU learning and transferred via fine-tuning to the other claim check-worthiness detection datasets, which are subsequently trained on as PU data. On top of vanilla PU learning, we introduce \textit{Positive Unlabelled Conversion (PUC)} which relabels examples that are most confidently positive in the unlabelled data. A formal task definition, description of PU learning, and explanation of the \textit{PUC} extension are given in the following sections.

\subsection{Task Definition}
The fundamental task is binary text classification. In the case of positive-negative (PN) data, we have a labelled dataset $\mathcal{D}: \{(x, y)\}$ with input features $x \in \mathbb{R}^{d}$ and labels $y \in \{0, 1\}$. The goal is to learn a classifier $g: x \rightarrow (0,1)$ indicating the probability that the input belongs to the positive class. With PU data, the dataset $\mathcal{D}$ instead consists of samples $\{(x, s)\}$, where the value $s \in \{0,1\}$ indicates if a sample is labelled or not. The primary difference from the PN case is that, unlike for the labels $y$, a value of $s = 0$ does not denote the sample is negative, but that the label is unknown. The goal is then to learn a PN classifier $g$ using a PU classifier $f: x \rightarrow (0,1)$ which predicts whether or not a sample is labelled~\cite{elkan2008learning}. 

\subsection{PU Learning}
\label{sec:pu_learning}
Our overall approach is depicted in~\autoref{fig:puc}. We begin with an explanation of the PU learning algorithm described in~\cite{elkan2008learning}. Assume that we have a dataset randomly drawn from some probability distribution $p(x,y,s)$, where samples are of the form $(x,s), ~s \in \{0,1\}$ and $s=1$ indicates that the sample is labelled. The variable $y$ is unknown, but we make two assumptions which allow us to derive an estimator for probabilities involving $y$. The first is that:
\begin{equation}
    p(y = 0 | s = 1) = 0
\end{equation}
In other words, if we know that a sample is labelled, then that label cannot be 0. The second assumption is that labelled samples are Selected Completely At Random from the underlying distribution (also known as the SCAR assumption). Check-worthiness data can be seen as an instance of SCAR PU data; annotators tend to only label those instances which are very clearly check-worthy in \textit{their} opinion~\cite{konstantinovskiy2018towards}. When combined across several annotators, we assume this leads to a random sample from the total set of check-worthy statements.

Given this, a classifier $f : x \rightarrow (0,1)$ is trained to predict $p(s=1|x)$ from the PU data. It is then employed to train a classifier $g$ to predict $p(y=1|x)$ by first estimating $c = p(s=1|y=1)$ on a set of validation data. Considering a validation set $V$ where $P \subset V$ is the set of positive samples in $V$, $c$ is estimated as:
\begin{equation}
\label{eq:c_estimate}
    c \approx \frac{1}{|P|}\sum_{x \in P}f(x)
\end{equation}
This says our estimate of $p(s=1|y=1)$ is the average confidence of our classifier on known positive samples. Next, we can estimate $E_{p(x,y,s)}[h(x,y)]$ for any arbitrary function $h$ empirically from a dataset of $k$ samples as follows:
\begin{dmath}
    E[h] = \frac{1}{k}(\sum_{(x,s=1)}h(x,1) + \sum_{(x,s=0)}w(x)h(x,1) + (1-w(x))h(x,0))
\end{dmath}
\begin{align}
    w(x) &= p(y=1|x,s=0) \notag\\ &= \frac{1-c}{c}\frac{p(s=1|x)}{1-p(s=1|x)}
\end{align}
In this case, $c$ is estimated using~\autoref{eq:c_estimate} and $p(s=1|x)$ is estimated using the classifier $f$. The derivations for these equations can be found in~\cite{elkan2008learning}.

To estimate $p(y=1|x)$ empirically, the unlabelled samples in the training data are duplicated, with one copy negatively labelled and one copy positively labelled. Each copy is trained on with a weighted loss $w(x)$ when the label is positive and $1 - w(x)$ when the label is negative. Labelled samples are trained on normally (i.e. a single copy with unit weight).

\subsection{Positive Unlabelled Conversion}
For \textit{PUC}, the motivation is to relabel those samples from the unlabelled data which are very clear cut positive. To accomplish this, we start with the fact that one can also estimate the prior probability of a sample having a positive label using $f$. If instead of $h$ we want to estimate $E[y] = p(y=1)$, the following is obtained:
\begin{equation}
    p(y=1) \approx \frac{1}{k}(\sum_{x,s=1}1 + \sum_{x,s=0}w(x))
\end{equation}
This estimate is then utilized to convert the most confident unlabelled samples into positives. First, all of the unlabelled samples are ranked according to their calculated weight $w(x)$. The ranked samples are then iterated through and converted into positive-only samples until the distribution of positive samples is greater than or equal to the estimate of $p(y=1)$. Unlike in vanilla PU learning, these samples are discretized to have a positive weight of 1, and trained on by the classifier $g$ once per epoch as positive samples along with the labelled samples. The remaining unlabelled data are trained on in the same way as in vanilla PU learning. 

\subsection{Implementation}
In order to create a unified approach to check-worthiness detection, transfer learning from Wikipedia citation needed detection is employed. 
To accomplish this, we start with a training dataset $\mathcal{D}^{s}$ of statements from Wikipedia featured articles that are either labelled as containing a citation (positive) or unlabelled. We train a classifier $f^{s}$ on this dataset and obtain a classifier $g^{s}$ via \textit{PUC}. For comparison, we also train models with vanilla PU learning and PN learning as baselines. The network architecture for both $f^{s}$ and $g^{s}$ is BERT~\cite{devlin2019bert}, a large pretrained transformer-based~\cite{vaswani2017attention} language model. We use the HuggingFace transformers implementation of the 12-layer 768 dimensional variation of BERT~\cite{Wolf2019HuggingFacesTS}. The classifier in this implementation is a two layer neural network acting on the \texttt{[CLS]} token.

From $g^{s}$, we train a classifier $g^{t}$ using downstream check-worthiness detection dataset $D^{t}$ by initializing $g^{t}$ with the base BERT network from $g^{s}$ and using a new randomly initialized final layer. In addition, we train a model $f^{t}$ on the target dataset, and train $g^{t}$ with \textit{PUC} from this model to obtain the final classifier. As a baseline, we also experiment with training on just the dataset $D^{t}$ without any pretraining. In the case of citation needed detection, since the data comes from the same domain we simply test on the test split of statements labelled as ``citation needed'' using the classifier $g^{s}$. We compare our models to the published state of the art baselines on each dataset.

For all of our models ($f^s$, $g^s$, $f^t$, $g^t$) we train for two epochs, saving the weights with the best F1 score on validation data as the final model. Training is performed with a max learning rate of 3e-5 and a triangular learning rate schedule~\cite{howard2018universal} that linearly warms up for 200 training steps, then linearly decays to 0 for the rest of training. For regularization we add L2 loss with a coefficient of 0.01, and dropout with a rate of 0.1. Finally, we split the training sets into 80\% train and 20\% validation, and train with a batch size of 8. 
The code to reproduce our experiments can be found here.\footnote{\url{https://github.com/copenlu/check-worthiness-pu-learning}} 

\section{Experimental Results}

\begin{table*}[t]
    \centering
    \fontsize{10}{10}\selectfont
    \begin{tabular}{l c c c : c c c}
    \toprule 
    Method & P & R & \multicolumn{1}{c}{F1} & eP & eR & eF1\\
    \midrule 
       \citealt{redi2019citation}  & 75.3& 70.9& 73.0 [76.0]*& - & - & -\\
    \hdashline
       \rule{0pt}{2ex}BERT  & \underline{78.8 $\pm$ 1.3}& 83.7 $\pm$ 4.5& 81.0 $\pm$ 1.5 & 79.0 & 85.3 & 82.0 \\
       BERT + PU  & \textbf{78.8 $\pm$ 0.9}& \underline{84.3 $\pm$ 3.0}& \underline{81.4 $\pm$ 1.0} & 79.0 & \underline{85.6} & \underline{82.2}\\
       BERT + \textit{PUC}  & 78.4 $\pm$ 0.9& \textbf{85.6 $\pm$ 3.2}& \textbf{81.8 $\pm$ 1.0} & 78.6 & \textbf{87.1} & \textbf{82.6}\\
    \bottomrule 

    \end{tabular}
    \caption{F1 and ensembled F1 score for citation needed detection training on the FA split and testing on the LQN split of \cite{redi2019citation}. The FA split contains statements with citations from featured articles and the LQN split consists of statements which were flagged as not having a citation but needing one. Listed are the mean, standard deviation, and ensembled results across 15 seeds (eP, eR, and eF1). \textbf{Bold} indicates best performance, \underline{underline} indicates second best. *The reported value is from rerunning their released model on the test dataset. The value in brackets is the value reported in the original paper.}
    \label{tab:citation_detection_results}
\end{table*}
To what degree is claim check-worthiness detection a PU learning problem, and does this enable a unified approach to check-worthiness detection? In our experiments, we progressively answer this question by answering the following: 1) is PU learning beneficial for the tasks considered? 2) Does PU citation needed detection transfer to rumour detection? 3) Does PU citation needed detection transfer to political speeches? To investigate how well the data in each domain reflects the definition of a check-worthy statement as one which ``makes an assertion about the world which is checkable'' and thus understand subjectivity in the annotations, we perform a dataset analysis comparing the provided labels of the top ranked check-worthy claims from the \textit{PUC} model with the labels given by two human annotators. In all experiments, we report the mean performance of our models and standard deviation across 15 different random seeds. Additionally, we report the performance of each model ensembled across the 15 runs through majority vote on each sample.

\subsection{Datasets\footnote{See supplemental material for links to datasets}}
\paragraph{Wikipedia Citations}
We use the dataset from \citet{redi2019citation} for citation needed detection. The dataset is split into three sets: one coming from featured articles (deemed `high quality', 10k positive and 10k negative statments), one of statements which have no citation but have been flagged as needing one (10k positive, 10k negative), and one of statements from random articles which have citations (50k positive, 50k negative). In our experiments the models were trained on the high quality statements from featured articles and tested on the statements which were flagged as `citation needed'. The key differentiating features of this dataset from the other two datasets are: 1) the domain of text is Wikipedia and 2) annotations are based on the decisions of Wikipedia editors following Wikipedia guidelines for citing sources\footnote{\url{https://en.wikipedia.org/wiki/Wikipedia:Citing_sources}}.

\paragraph{Twitter Rumours}
The PHEME dataset of rumours is employed for Twitter claim check-worthiness detection~\cite{zubiaga2016analysing}. The data consists of 5,802 annotated tweets from 5 different events, where each tweet is labelled as rumourous or non-rumourous (1,972 rumours, 3,830 non-rumours). We followed the leave-one-out evaluation scheme of~\cite{zubiaga2017exploiting}, namely, we performed a 5-fold cross-validation for all methods, training on 4 events and testing on 1. The key differentiating features of this dataset from the other two datasets are: 1) the domain of data is tweets and 2) annotations are collected from professional journalists specifically for building a dataset to train machine learning models.

\paragraph{Political Speeches}
The dataset we adopted in the political speeches domain is the same as in~\citet{hansen2019neural}, consisting of 4 political speeches from the 2018 Clef CheckThat! competition~\cite{atanasova2018overview} and 3 political speeches from ClaimRank~\cite{jaradat2018claimrank} (2,602 statements total). We performed a 7-fold cross-validation, using 6 splits as training data and 1 as test in our experimental setup. The data from ClaimRank is annotated using the judgements from 9 fact checking organizations, and the data from Clef 2018 is annotated by factcheck.org. The key differentiating features of this dataset from the other two datasets are: 1) the domain of data is transcribed spoken utterances from political speeches and 2) annotations are taken from 9 fact checking organizations gathered independently.

\subsection{Is PU Learning Beneficial for Citation Needed Detection?}
\begin{table*}[ht]
    \centering
    \fontsize{10}{10}\selectfont
    \begin{tabular}{l c c c : c c c}
    \toprule
    Method & $\mu$P & $\mu$R & \multicolumn{1}{c}{$\mu$F1}& eP & eR & eF1\\
    \midrule
       \citealt{zubiaga2017exploiting}  & 66.7& 55.6& 60.7 & - & - & -\\
       BiLSTM & 62.3 &	56.4 &	59.0 & - & - & -\\
    \hdashline
       \rule{0pt}{2ex}BERT  & \underline{69.9 $\pm$ 1.7}& 60.8 $\pm$ 2.6& 65.0 $\pm$ 1.3 & 71.3 & 61.9	& 66.3\\
       BERT + Wiki  & 69.3 $\pm$ 1.6& 61.4 $\pm$ 2.6& 65.1 $\pm$ 1.2 & 70.7 & 62.2 & 66.2\\
       BERT + WikiPU  & \underline{69.9 $\pm$ 1.3}& 62.5 $\pm$ 1.6& 66.0 $\pm$ 1.1 & \textbf{72.2} & 64.6 & 68.2\\
       BERT + Wiki\textit{PUC}  & \textbf{70.1 $\pm$ 1.1}& 61.8 $\pm$ 1.8& 65.7 $\pm$ 1.0 & \underline{71.5} & 62.7 & 66.8\\
       BERT + PU  & 68.7 $\pm$ 1.2& 64.7 $\pm$ 1.8& 66.6 $\pm$ 0.9 & 69.9 & 65.2 & 67.5\\
       BERT + \textit{PUC}  & 68.1 $\pm$ 1.5& 65.3 $\pm$ 1.6& 66.6 $\pm$ 0.9 & 69.1 & 66.3 & 67.7\\
       BERT + PU + WikiPU  & 68.4 $\pm$ 1.2& \textbf{66.1 $\pm$ 1.2}& \textbf{67.2 $\pm$ 0.6} & 69.3 & \underline{67.2} & \underline{68.3}\\
       BERT + \textit{PUC} + WikiPUC  & 68.0 $\pm$ 1.4& \underline{66.0 $\pm$ 2.0}& \underline{67.0 $\pm$ 1.3} & 69.4 & \textbf{67.5} & \textbf{68.5}\\
    \bottomrule

    \end{tabular}
    \caption{micro-F1 ($\mu$F1) and ensembled F1 (eF1) performance of each system on the PHEME dataset. Performance is averaged across the five splits of~\cite{zubiaga2017exploiting}. Results show the mean, standard deviation, and ensembled score across 15 seeds. \textbf{Bold} indicates best performance, \underline{underline} indicates second best.}
    \label{tab:pheme_results}
\end{table*}


Our results for citation needed detection are given in \autoref{tab:citation_detection_results}. The vanilla BERT model already significantly outperforms the state of the art model from Redi et al.~\shortcite{redi2019citation} (a GRU network with global attention) by 6 F1 points. We see further gains in performance with PU learning, as well as when using \textit{PUC}. Additionally, the models using PU learning have lower variance, indicating more consistent performance across runs. The best performing model we see is the one trained using \textit{PUC} with an F1 score of 82.6. We find that this confirms our hypothesis that citation data is better seen as a set of positive and unlabelled data when used for check-worthiness detection. In addition, it gives some indication that PU learning improves the generalization power of the model, which could make it better suited for downstream tasks.


\subsection{Does PU Citation Needed Detection Transfer to Rumour Detection?}
\subsubsection{Baselines}
The best published method that we compare to is the CRF from~\cite{zubiaga2017exploiting}. which utilizes a combination of content and social features. Content features include word vectors, part-of-speech tags, and various lexical features, and social features include tweet count, listed count, follow ratio, age, and whether or not a user is verified. The CRF acts on a timeline of tweets, making it contextual. In addition, we include results from a 2-layer BiLSTM with FastText embeddings~\cite{bojanowski2017enriching}. There exist other deep learning models which have been developed for this task, including \cite{ma2019detect} and \cite{abulaish2019graph}, but they do not publish results on the standard splits of the data and we were unable to recreate their results, and thus are omitted.

\subsubsection{Results}
The results for the tested systems are given in \autoref{tab:pheme_results}. Again we see large gains from BERT based models over the baseline from \cite{zubiaga2017exploiting} and the 2-layer BiLSTM. Compared to training solely on PHEME, fine tuning from basic citation needed detection sees little improvement (0.1 F1 points). However, fine tuning a model trained using PU learning leads to an increase of 1 F1 point over the non-PU learning model, indicating that PU learning enables the Wikipedia data to be useful for transferring to rumour detection i.e. the improvement is not only from a better semantic representation learned from Wikipedia data. For \textit{PUC}, we see an improvement of 0.7 F1 points over the baseline and lower overall variance than vanilla PU learning, meaning that the results with \textit{PUC} are more consistent across runs. The best performing models also use PU learning on in-domain data, with the best average performance being from the models trained using PU/\textit{PUC} on in domain data and initialized with weights from a Wikipedia model trained using PU/\textit{PUC}. When models are ensembled, pretraining with vanilla PU learning improves over no pretraining by almost 2 F1 points, and the best performing models which are also trained using PU learning on in domain data improve over the baseline by over 2 F1 points. We conclude that framing rumour detection on Twitter as a PU learning problem leads to improved performance.




Based on these results, we are able to confirm two of our hypotheses. The first is that Wikipedia citation needed detection and rumour detection on Twitter are indeed similar tasks, and a unified approach for both of them is possible. Pretraining a model on Wikipedia provides a clear downstream benefit when fine-tuning on Twitter data, \textit{precisely when PU/PUC is used}. Additionally, training using \textit{PUC} on in domain Twitter data provides further benefit. This shows that \textit{PUC} constitutes a unified approach to these two tasks.

The second hypothesis we confirm is that both Twitter and Wikipedia data are better seen as positive and unlabelled for claim check-worthiness detection. When pretraining with the data as a traditional PN dataset there is no performance gain and in fact a performance loss when the models are ensembled. PU learning allows the model to learn better representations for general claim check-worthiness detection.

To explain why this method performs better, \autoref{tab:citation_detection_results} and \autoref{tab:pheme_results} show that \textit{PUC} improves model recall at very little cost to precision. The aim of this is to mitigate the issue of subjectivity in the annotations of check-worthiness detection datasets noted in previous work \cite{konstantinovskiy2018towards}. Some of the effects of this are illustrated in \autoref{tab:pheme_pos_better} and \autoref{tab:pheme_neg_better} in \autoref{sec:puc_examples} The \textit{PUC} models are better at distinguishing rumours which involve claims of fact about people i.e. things that people said or did, or qualities about people. For non-rumours, the \textit{PUC} pretrained model is better at recognizing statements which describe qualitative information surrounding the events and information that is self-evident e.g. a tweet showing the map where the Charlie Hebdo attack took place.

\subsection{Does PU Citation Needed Detection Transfer to Political Speeches?}
\subsubsection{Baselines}
The baselines we compare to are the state of the art models from ~\citet{hansen2019neural} and \citet{konstantinovskiy2018towards}. The model from \citet{konstantinovskiy2018towards} consists of InferSent embeddings~\cite{conneau2017supervised} concatenated with POS tag and NER features passed through a logistic regression classifier. The model from \citet{hansen2019neural} is a bidirectional GRU network acting on syntatic parse features concatenated with word embeddings as the input representation.

\subsubsection{Results}
The results for political speech check-worthiness detection are given in \autoref{tab:clef_results}. We find that the BERT model initialized with weights from a model trained on plain Wikipedia citation needed statements performs the best of all models. As we add transfer learning and PU learning, the performance steadily drops. We perform a dataset analysis to gain some insight into this effect in \S\ref{sec:dataset_analysis}.
\begin{table}
    \centering
    \fontsize{10}{10}\selectfont
    \begin{tabular}{l c}
    \toprule
    Method & MAP\\
    \midrule
        \citealt{konstantinovskiy2018towards} & 26.7\\
       \citealt{hansen2019neural}  & 30.2\\
    \hdashline
       \rule{0pt}{2ex}BERT  &33.0 $\pm$ 1.8\\
       BERT + Wiki  & \textbf{34.4 $\pm$ 2.7}\\
       BERT + WikiPU  & \underline{33.2 $\pm$ 1.7}\\
       BERT + Wiki\textit{PUC}  & 31.7 $\pm$ 1.8\\
       BERT + PU  & 18.8 $\pm$ 3.7\\
       BERT + \textit{PUC}  & 26.7 $\pm$ 2.8\\
       BERT + PU + WikiPU  & 16.8 $\pm$ 3.5\\
       BERT + \textit{PUC} + Wiki\textit{PUC}  & 27.8 $\pm$ 2.7\\
    \bottomrule

    \end{tabular}
    \caption{Mean average precision (MAP) of models on political speeches. \textbf{Bold} indicates best performance, \underline{underline} indicates second best.}
    \label{tab:clef_results}
\end{table}

\subsection{Dataset Analysis}
\label{sec:dataset_analysis}
In order to understand our results in the context of the selected datasets, we perform an analysis to learn to what extent the positive samples in each dataset reflect the definition of a check-worthy claim as ``an assertion about the world that is checkable''. We ranked all of the statements based on the predictions of 15 \textit{PUC} models trained with different seeds, where more positive class predictions means a higher rank (thus more check-worthy), and had two experts manually relabel the top 100 statements. The experts were informed to label the statements based on the definition of check-worthy given above. We then compared the manual annotation to the original labels using F1 score. Higher F1 score indicates the dataset better reflects the definition of check-worthy we adopt in this work. Our results are given in \autoref{tab:relabel_results}.
\begin{table}
    \centering
    \fontsize{10}{10}\selectfont
    \begin{tabular}{l c c c}
    \toprule
    Dataset & P & R & F1\\
    \midrule
                  &  81.7 &87.0 &84.3\\
       Wikipedia  & 84.8 & 87.0 & 85.9\\
                  & \textit{83.3}& \textit{87.0}& \textit{85.1}\\
       \hdashline
        \rule{0pt}{2ex}  & 87.5& 82.4& 84.8 \\
       Twitter  & 86.3 & 81.2 & 83.6\\
                & \textit{86.9}& \textit{81.8}& \textit{84.2} \\
       \hdashline
        \rule{0pt}{2ex} &33.8& 89.3& 49.0\\
       Politics  &31.1&  100.0&  47.5\\
                &\textit{32.5} &\textit{94.7}& \textit{48.3}\\
    \bottomrule

    \end{tabular}
    \caption{F1 score comparing manual relabelling of the top 100 predictions by \textit{PUC} model with the original labels in each dataset by two different annotators. \textit{Italics} are average value between the two annotators.}
    \label{tab:relabel_results}
\end{table}

We find that the Wikipedia and Twitter datasets contain labels which are more general, evidenced by similar high F1 scores from both annotators ($>$ 80.0). For political speeches, we observe that the human annotators both found many more examples to be check-worthy than were labelled in the dataset. This is evidenced by examples such as \textit{It's why our unemployment rate is the lowest it's been in so many decades} being labelled as not check-worthy and \textit{New unemployment claims are near the lowest we've seen in almost half a century} being labelled as check-worthy in the same document in the dataset's original annotations. This characteristic has been noted for political debates data previously~\cite{konstantinovskiy2018towards}, which was also collected using the judgements of independent fact checking organizations~\cite{gencheva-etal-2017-context-fixed}. Labels for this dataset were collected from various news outlets and fact checking organizations, which may only be interested in certain types of claims such as those most likely to be false. This makes it difficult to train supervised machine learning models for general check-worthiness detection based solely on text content and document context due to labelling inconsistencies. 

\section{Discussion and Conclusion}
In this work, we approached claim check-worthiness detection by examining how to unify three distinct lines of work. We found that check-worthiness detection is challenging in any domain as there exist stark differences in how annotators judge what is check-worthy. We showed that one can correct for this and improve check-worthiness detection across multiple domains by using positive unlabelled learning. Our method enabled us to perform a structured comparison of datasets in different domains, developing a unified approach which outperforms state of the art in 2 of 3 domains and illuminating to what extent these datasets reflect a general definition of check-worthy. 

Future work could explore different neural base architectures. Further, 
it could potentially benefit all tasks to consider the greater context in which statements are made. We would also like to acknowledge again that all experiments have only focused on English language datasets; developing models for other, especially low-resource languages, would likely result in additional challenges. We hope that this work will inspire future research on check-worthiness detection, which we see as an under-studied problem, with a focus on developing resources and models across many domains such as Twitter, news media, and spoken rhetoric.

\section*{Acknowledgements}
$\begin{array}{l}\includegraphics[width=1cm]{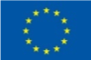} \end{array}$ This project has received funding from the European Union's Horizon 2020 research and innovation programme under the Marie Sk\l{}odowska-Curie grant agreement No 801199.

\bibliography{anthology,acl2020}
\bibliographystyle{acl_natbib}

\newpage
\appendix

\section{Examples of PUC Improvements for Rumour Detection}
\label{sec:puc_examples}

Examples of improvements for rumour detection using \textit{PUC} can be found in \autoref{tab:pheme_pos_better}.
\begin{table*}[h]
    \centering
    \begin{tabular}{p{12cm} c c}
    \toprule
    \multicolumn{1}{c}{Rumour text} & nPUC & \multicolumn{1}{c}{nBaseline} \\
    \midrule
       Germanwings co-pilot had serious depressive episode: Bild newspaper http://t.co/RgSTrehD21 & 13 & 5\\
       \hline
       Now hearing 148 passengers + crew on board the \#A320 that has crashed in southern French Alps. \#GermanWings flight. @BBCWorld & 10 & 2 \\
       \hline
       It appears that \#Ferguson PD are trying to assassinate Mike Brown's character after literally assassinating Mike Brown. & 13 & 5 \\
       \hline
       \#Ferguson cops beat innocent man then charged him for bleeding on them: http://t.co/u1ot9Eh5Cq via @MichaelDalynyc http://t.co/AGJW2Pid1r & 9 & 2\\
    \bottomrule

    \end{tabular}
    \caption{Examples of rumours which the \textit{PUC} model judges correctly vs the baseline model with no pretraining on citation needed detection. n* is the number of models among the 15 seeds which predicted the correct label (rumour).}
    \label{tab:pheme_pos_better}
\end{table*}
\begin{table*}
    \centering
    \begin{tabular}{p{12cm} c c}
    \toprule
    \multicolumn{1}{c}{Non-Rumour text} & nPUC & \multicolumn{1}{c}{nBaseline} \\
    \midrule
       A female hostage stands by the front entrance of the cafe as she turns the lights off in Sydney. \#sydneysiege http://t.co/qNfCMv9yZt & 11 & 5\\
       \hline
       Map shows where gun attack on satirical magazine \#CharlieHebdo took place in central Paris http://t.co/5AZAKumpNd http://t.co/ECFYztMVk9 & 10 & 4 \\
       \hline
       "Hands up! Don't shoot!" \#ferguson https://t.co/svCE1S0Zek & 12 & 7 \\
       \hline
       Australian PM Abbott: Motivation of perpetrator in Sydney hostage situation is not yet known - @9NewsAUS http://t.co/SI01B997xf & 10 & 6\\
    \bottomrule

    \end{tabular}
    \caption{Examples of non-rumours which the \textit{PUC} model judges correctly vs the baseline model with no pretraining on citation needed detection. n* is the number of models among the 15 seeds which predicted the correct label (non-rumour).}
     \label{tab:pheme_neg_better}
\end{table*}

\section{Reproducibility}
\subsection{Computing Infrastructure}
All experiments were run on a shared cluster. Requested jobs consisted of 16GB of RAM and 4 Intel Xeon Silver 4110 CPUs. We used a single NVIDIA Titan X GPU with 12GB of RAM.

\subsection{Average Runtimes}
See \autoref{tab:runtimes} for model runtimes.
\begin{table*}
    \centering
    \fontsize{10}{10}\selectfont
    \begin{tabular}{l c c c}
    \toprule
    Method & Wikipedia & PHEME & Political Speeches\\
    \midrule
       \rule{0pt}{2ex}BERT&  34m30s& 14m25s& 8m11s\\
       BERT + PU& 40m7s& 20m40s&  15m38s\\
       BERT + \textit{PUC}& 40m8s& 21m20s& 15m32s\\
       BERT + Wiki& - & 14m28s& 8m50s\\
       BERT + WikiPU& -& 14m25s& 8m41s\\
       BERT + Wiki\textit{PUC}& -& 14m28s& 8m38s\\
       BERT + PU + WikiPU& -& 20m41s& 15m32s\\
       BERT + \textit{PUC} + WikiPUC& -& 21m52s& 15m40s\\
    \bottomrule

    \end{tabular}
    \caption{Average runtime of each tested system for each split of the data}
    \label{tab:runtimes}
\end{table*}

\subsection{Number of Parameters per Model}
We used BERT with a classifier on top for each model which consists of 109,483,778 parameters.

\subsection{Validation Performance}
Validation performances for the tested models are given in \autoref{tab:validation}.
\begin{table*}
    \centering
    \fontsize{10}{10}\selectfont
    \begin{tabular}{l c c c}
    \toprule
    Method & Wikipedia & PHEME & Political Speeches\\
    \midrule
       \rule{0pt}{2ex}BERT&  88.9& 81.6& 31.3\\
       BERT + PU& 89.0& 83.7& 18.2\\
       BERT + \textit{PUC}& 89.2& 82.8& 32.0\\
       BERT + Wiki& - & 80.8& 32.3\\
       BERT + WikiPU& -& 82.0& 35.7\\
       BERT + Wiki\textit{PUC}& -& 80.4& 34.3\\
       BERT + PU + WikiPU& -& 82.9& 33.3\\
       BERT + \textit{PUC} + WikiPUC& -& 84.1& 34.0\\
    \bottomrule

    \end{tabular}
    \caption{Validation F1 performances for each tested model.}
    \label{tab:validation}
\end{table*}

\subsection{Evaluation Metrics}
The primary evaluation metric used was F1 score. We used the sklearn implementation of \texttt{precision\_recall\_fscore\_support}, which can be found here: \url{https://scikit-learn.org/stable/modules/generated/sklearn.metrics.precision_recall_fscore_support.html}. Briefly:
\begin{equation*}
   p = \frac{tp}{tp + fp} 
\end{equation*}
\begin{equation*}
   r = \frac{tp}{tp + fn} 
\end{equation*}
\begin{equation*}
   F1 = \frac{2*p*r}{p+r} 
\end{equation*}
where $tp$ are true positives, $fp$ are false positives, and $fn$ are false negatives.

Additionally, we used the mean average precision calculation from the Clef19 Check That! challenge for political speech data, which can be found here: \url{https://github.com/apepa/clef2019-factchecking-task1/tree/master/scorer} Briefly:
\begin{equation*}
    \text{AP} = \frac{1}{|P|}\sum_{i}\frac{tp(i)}{i}
\end{equation*}
\begin{equation*}
    \text{mAP} = \frac{1}{|Q|}\sum_{q\in Q}\text{AP}(q)
\end{equation*}
where $P$ are the set of positive instances, $tp(i)$ is an indicator function which equals one when the $i$th ranked sample is a true positive, and $Q$ is the set of queries. In this work $Q$ consists of the ranking of statements from each split of the political speech data.

\subsection{Links to Data}
\begin{itemize}
    \item Citation Needed Detection~\cite{redi2019citation}:  \url{https://drive.google.com/drive/folders/1zG6orf0_h2jYBvGvso1pSy3ikbNiW0xJ}
    
    \item PHEME~\cite{zubiaga2016analysing}: \url{https://figshare.com/articles/PHEME_dataset_for_Rumour_Detection_and_Veracity_Classification/6392078}.
    
    \item Political Speeches: We use the same 7 splits as used in~\cite{hansen2019neural}. The first 5 can be found here: \url{http://alt.qcri.org/clef2018-factcheck/data/uploads/clef18_fact_checking_lab_submissions_and_scores_and_combinations.zip}. The files can be found under "task1\_test\_set/English/task1-en-file(3,4,5,6,7)". The last two files can be found here: \url{https://github.com/apepa/claim-rank/tree/master/data/transcripts_all_sources}. The files are ``clinton\_acceptance\_speech\_ann.tsv'' and ``trump\_inauguration
    \_ann.tsv''.
\end{itemize}

\subsection{Hyperparameters}
We found that good defaults worked well, and thus did not perform hyperparameter search. The hyperparameters we used are given in \autoref{tab:hyperparams}.

\begin{table}
    \centering
    \fontsize{10}{10}\selectfont
    \begin{tabular}{l c}
    \toprule
    Hyperparameter & Value \\
    \midrule
      Learning Rate& 3e-5 \\
      Weight Decay& 0.01 \\
      Batch Size& 8 \\
      Dropout& 0.1 \\
      Warmup Steps& 200 \\
      Epochs& 2 \\
    \bottomrule

    \end{tabular}
    \caption{Validation F1 performances used for each tested model.}
    \label{tab:hyperparams}
\end{table}

\end{document}